\documentclass[conference]{IEEEtran}
\IEEEoverridecommandlockouts
\usepackage{cite}
\usepackage{amsmath,amssymb,amsfonts}
\usepackage{algorithmic}
\usepackage{graphicx}
\usepackage{textcomp}
\usepackage{xcolor}
\newcommand\mdoubleplus{\mathbin{+\mkern-10mu+}}

\def\BibTeX{{\rm B\kern-.05em{\sc i\kern-.025em b}\kern-.08em
    T\kern-.1667em\lower.7ex\hbox{E}\kern-.125emX}}
    
\usepackage{tikz}

\newcommand\copyrighttext{
  \footnotesize \textcopyright 2021 IEEE. Personal use of this material is permitted. Permission from IEEE must be obtained for all other uses, in any current or future media, including reprinting/republishing this material for advertising or promotional purposes, creating new collective works, for resale or redistribution to servers or lists, or reuse of any copyrighted component of this work in other works. Accepted to be Published in: Proceedings of the IJCNN 2021: International Joint Conference on Neural Networks, Padua, Italy 18-23 July 2021.}
\newcommand\copyrightnotice{
\begin{tikzpicture}[remember picture,overlay]
\node[anchor=south,yshift=10pt] at (current page.south) {\fbox{\parbox{\dimexpr\textwidth-\fboxsep-\fboxrule\relax}{\copyrighttext}}};
\end{tikzpicture}
}

\begin{document}

\title{RNN-BOF: A Multivariate Global Recurrent Neural Network for Binary Outcome Forecasting of Inpatient Aggression\\
\thanks{This research was supported by the Australian Research Council under grant DE190100045.}
}

\author{\IEEEauthorblockN{Aidan Quinn\IEEEauthorrefmark{1}, Melanie Simmons\IEEEauthorrefmark{2}, Benjamin Spivak\IEEEauthorrefmark{2} and Christoph Bergmeir\IEEEauthorrefmark{1}}
\IEEEauthorblockA{\IEEEauthorrefmark{1}Faculty of Information Technology\\
Monash University, Melbourne, Australia\\
Email: aidanpcquinn@gmail.com, christoph.bergmeir@monash.edu}
\IEEEauthorblockA{\IEEEauthorrefmark{2}Department of Psychology\\
Swinburne University of Technology, Melbourne, Australia\\
Email: \{msimmons, bspivak\}@swin.edu.au}}

\maketitle
\copyrightnotice

\begin{abstract}
Psychometric assessment instruments aid clinicians by providing methods of assessing the future risk of adverse events such as aggression. Existing machine learning approaches have treated this as a classification problem, predicting the probability of an adverse event in a fixed future time period from the scores produced by both psychometric instruments and clinical and demographic covariates. We instead propose modelling a patient's future risk using a time series methodology that learns from longitudinal data and produces a probabilistic binary forecast that indicates the presence of the adverse event in the next time period. Based on the recent success of Deep Neural Nets for globally forecasting across many time series, we introduce a global multivariate Recurrent Neural Network for Binary Outcome Forecasting, that trains from and for a population of patient time series to produce individual probabilistic risk assessments. We use a moving window training scheme on a real world dataset of 83 patients, where the main binary time series represents the presence of aggressive events and covariate time series represent clinical or demographic features and psychometric measures. On this dataset our approach was capable of a significant performance increase against both benchmark psychometric instruments and previously used machine learning methodologies.
\end{abstract}


\section{Introduction}
Psychometric risk assessment instruments offer a structured measure that can aid clinicians in management of their patients by indicating the future risk of an adverse event such as aggression or violence. Tests such as the Dynamic Appraisal of Situational Aggression (DASA) \cite{OgloffDASA} and Short-Term Assessment of Risk and Treatability (START) \cite{STARTOriginal} produce scores on an ordinal scale based on the presence of multiple risk domains, which in turn correspond to the likelihood of an adverse event occurring. Probabilistic machine learning classifiers have been built from these instruments and additional clinical and demographic covariates \cite{IsaacQuantitative,ParghiAssessing,OrruMachine}, although this approach has no explicit way of learning temporal relationships that may occur between recurrent assessments or in the history of adverse events. As such, we propose a methodology for modelling this type of risk assessment as a probabilistic global multivariable time series forecasting task.

A local multivariable forecasting approach would produce a per-patient model from a binary time series representing the adverse event and covariate series representing clinical and demographic features. Instead, we train a single global forecaster for a population of patients allowing for cross learning between patients, more complex modelling methods, and no requirement for homogeneity between patients' time series \cite{MonteroPrinciples} or an instance of the adverse event within the patient's history. For this task, we propose a sliding window time series based modelling scheme and a Recurrent Neural Network architecture for Binary Outcome Forecasting, RNN-BOF. An implementation of our method is available online\footnote[1]{Implementation available at https://github.com/aidanpcquinn/RNNBOF}. Unlike other recent global multivariable forecasters \cite{SenThink,LiCombining}, our model is capable of producing probabilities from a main binary time series, allowing us to build a generalised and time aware risk assessment tool.

RNN-BOF uses stacked Long-Short Term Memory cells (LSTM) constructed as a many-to-one forecaster with a single neuron sigmoid activation output layer, an adaptation from other RNN forecasters \cite{BandaraSales} to allow for the probabilistic and binary day-ahead outcome prediction. The model assigns any given patient a probability associated with the presence of the adverse event occurring in the next time interval from fixed length input windows. We compare this approach on a real world dataset against both the psychometric instruments themselves, and the state-of-art machine learning classifiers as used for clinical risk assessment.

We evaluate RNN-BOF on the Thomas Embling Hospital risk assessment dataset (TEH dataset) which contains daily recordings indicating the presence of an aggressive event, daily psychometric measures using the START and DASA instruments, and other static and dynamic clinical and demographic features. We then evaluate forecasting performance using the Precision Recall Gain curve (PRG) \cite{PRG}, a scoring rule that we propose is best suited to assess a model's performance.

The main contributions of our work are as follows: \textbf{(A)} We present RNN-BOF, a novel implementation of an RNN as a scored global multivariable binary time series forecaster using a pooled sliding window training scheme. \textbf{(B)} We show experimental evidence that our methodology is capable of outperforming benchmark psychometric instruments and traditional machine learning models on the task of predicting the presence of inpatient aggression on a real world dataset. \textbf{(C)} Furthermore, our evidence suggests that longitudinal data which represents the presence of an event, alongside covariate data and categorical features, could benefit from being modelled as a global time series problem, even when a probability is required.

\section{Background and related works}

In this section we briefly describe related works in psychometric evaluation and time series forecasting.

\subsection{Psychometrics and machine learning}

We have identified two studies that directly employed psychometric instruments and clinical covariates alongside machine learning techniques. Galatzer-Levy et al.~\cite{IsaacQuantitative} tested the efficacy of various models including Support Vector Machines (SVM), Random Forests (RF) and kernel ridge regressions trained to predict the presence of a traumatic stress disorder diagnosis within 15 months of a traumatic event. The models were trained on patient data collected at the time of the event, general demographic data, physiological responses, and an assessment on the Acute Stress Disorder scale psychometric instrument. Their approach was able to outperform the benchmark predictive power of the scale itself. A similar work \cite{ParghiAssessing} assessed  high-risk psychiatric inpatients using the Suicide Crisis Inventory instrument which is composed of 49 five-point ordinal scales. They trained models including RFs and gradient boosted decision trees (GBDT) to predict the occurrence of suicide attempts over a one month period although they made no comparison to the baseline psychometric evaluation. 

\subsection{Global multivariable time series forecasting}

Statistical local forecasting methods have consistently out-performed neural networks and other machine learning techniques in forecasting competitions for decades~\cite{MakridakisThe}. This changed only relatively recently in the M4 forecasting competition, which was won by a hybrid model that employed both statistical and neural network forecasting, training models globally across all time series~\cite{SmylA}. Following this success and a subsequent rise in popularity, Montero-Manso and Hyndman~\cite{MonteroPrinciples} established theoretical bounds of effectiveness for global cross-series approaches as compared to local (per series) methods, suggested that the multivariable equivalent would inherit the beneficial properties of global cross-series methods, and offered empirical evidence for the lack of necessity of homogeneity across series. Hewamalage et al.~\cite{HewamalageGlobal,HewamalageRecurrent} supported these conclusions and identified RNNs with LSTM cells (RNN-LSTM) and Light Gradient Boosted Machines (LGBM) as strong performers on both real world and simulated datasets. 

However, to the best of our knowledge there is limited literature on multivariable time series forecasting where models are trained globally across many homogeneous multivariable datasets. The two multivariable global forecasters we have identified are DeepGLO \cite{SenThink} and TEDGE \cite{LiCombining}. Both of these models learn from a set of time series, where each series has covariate time series and static categorical features. Both use initial matrix factorisation to extract global trends; DeepGLO then uses a temporal convolution network to learn local patterns, while TEDGE makes use of a multivariable RNN-LSTM architecture. Either of these forecasters could suit our needs if not for the probabilistic or scored binary forecast requirement.

\subsection{Binary time series forecasters}

The majority of models for binary time series forecasting are from the auto-regressive family, with neural network approaches being scarce \cite{BautuEvolving}. gbARMA \cite{gbarma}, GLARMA \cite{glarma} and PC-ARIMA \cite{PC-ARIMA} are capable forecasters given a multivariate forecasting problem with a binary target series. However, they are all locally trained models \cite{SenThink}, meaning for a patient without any instances of the adverse event, the model would have no historical reference and therefore could not predict future instances. Bautu et al. \cite{BautuEvolving} used a probabilistic hyper-network to predict the direction of a stock price. Although theoretically globally trainable, the model's implementation is univariate, making it again unsuitable for our dataset.

\section{Methodology}

In this section we formally outline the learning task and compare this to previous formulations of similar tasks. We then describe the proposed model architecture, reasoning for the structure and components, details of the window training scheme, and the learning algorithm.

\subsection{Learning task} \label{Learning}

We first define the learning tasks as used in supervised learning classification works with similar feature sets \cite{IsaacQuantitative} and similar learning tasks \cite{ParghiAssessing} to the TEH dataset. For a population of patients, $p \in P$, we have the dataset with features $f \in \mathcal{F}$, where each $f$ can be binary, $f_b \in \{0,1\}$, the value of some bound ordinal scale $s$, $f_s \in \{1,2,... ,s_{max}\}$, or a value of continuous measurement, $f_c \in \mathbb{R}$. Examples of $f_b$ are general clinical data such as `history of drug use' or an outcome on a specific psychometric measure, such as `irritability' on the DASA assessment \cite{OgloffDASA}. Each $f_s$ could represent the summation of the binary measure components of an assessment such as DASA, or a score on a Likert scale \cite{likert}, while continuous measures, $f_c$, could refer to age or blood pressure at time of assessment. Each patient would have associated label, $y^p \in \{0,1\}$, representing the presence of a future event within a fixed future time frame. From this data we would learn a model that predicts a score or probability for the positive outcome (presence of the event) within the fixed time frame for any given patient, $\hat{y}_p \in \mathbb{R} | 0 \leq  \hat{y}_p \leq 1$.

The TEH dataset consists of daily longitudinal psychometric and clinical data with a feature set, $\mathcal{G}$, that exhibits the same feature types as those described in $\mathcal{F}$, alongside a daily recording representing the presence of an instance of aggression over the same daily periods. We propose modelling the future presence of aggression using a global multivariable binary time series forecaster, where the target series is a patient's binary time series indicating presence of aggression and associated covariate time series are features that vary over time. The task is, for any given patient, to forecast a score or probability associated with the presence of aggression in their next time interval. We note that the TEH dataset does not contain equal history length for each of the patients, contains static features that do not vary over time and is relatively small with only 83 patients total and a mean series length of 140 with a standard deviation of 106.

We formally define the new learning task as follows: 
For a patient $p \in P$, we represent the time series of the history of a dynamic feature, $d^{p,j}$ as the vector $\mathbf{d}^{p,j}_{1:Tp} = (d^{p,j}_1, d^{p,j}_2, ... , d^{p,j}_{Tp})$ where $Tp$ is the total series length of a patient, and $j$ is a dynamic feature $j \in \mathcal{F}_d$. We then define the set of all of a given patient's dynamic feature time series as $D^{p}_{1:Tp} = \{\mathbf{d}^{p,1}_{1:Tp}, \mathbf{d}^{p,2}_{1:Tp}, ...\}$. The static features for a patient, $s^{p,l}$, $l \in \mathcal{F}_s$, are represented as the set $S^p = \{s^{p,1}, s^{p,2}, ...\}$. The specific adverse event for a patient is represented by a binary vector $\mathbf{y}^{p}_{1:Tp} = (y^{p}_1, y^{p}_2, ... , y^{p}_{Tp})$, where $y_t^p \in \{0,1\}$ indicates the presence of the event. The complete dataset then contains the sets: $D = \{D^{p}_{1:Tp}\}^{p \in P}$, $S = \{S^{p}\}^{p \in I}$ and $Y = \{\mathbf{y}^{p}_{1:Tp}\}^{p \in P}$. The objective is to learn a model $M$ that takes a fixed length window of length $n \leq Tp$ from a given patient to produce a probability associated with the future positive patient outcome, namely $\hat{y}^p_{Tp+1} = M(D^{p}_{Tp-n:Tp}, S^{p},\mathbf{y}^{p}_{Tp-n:Tp}) $, where  $\hat{y}_{Tp+1} \in \mathbb{R} | 0 \leq  \hat{y}_{Tp+1} \leq 1$. We note that when only a length of $n = 1$ is considered by the model, this learning task matches the traditional future classification style approach with a bound of one day placed on the future time period of when the event can occur.

\subsection{RNN-BOF architecture}

\begin{figure*}[tb]
\centering
\includegraphics[width=\textwidth]{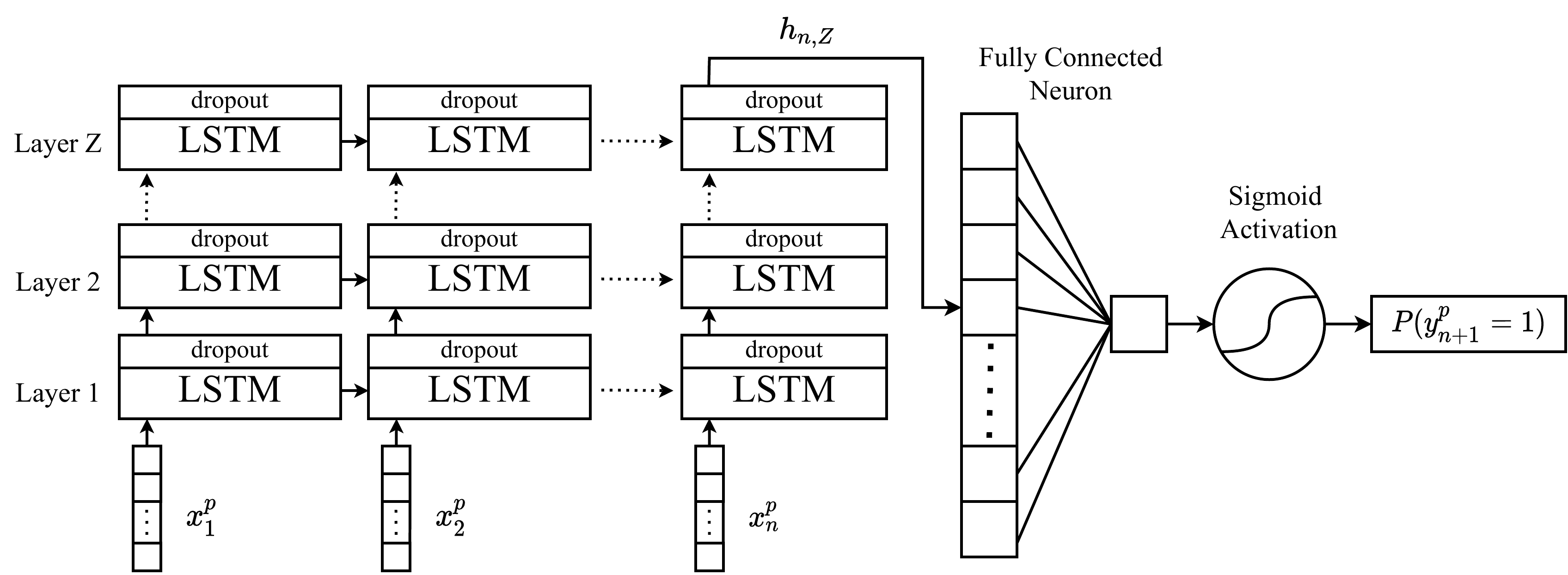}
\caption{An illustration of RNN-BOF's architecture with simplified LSTM cells.}
\label{fig:rnnlstm}
\end{figure*}

RNN-BOF uses a stacked Recurrent Neural Network architecture similar to that described by Bandara et al.~\cite{BandaraForecasting}, but with only the last LSTM cell's hidden state vector from the last stacked layer being fully connected to a single sigmoid output neuron for the single probabilistic binary forecast point (Fig. ~\ref{fig:rnnlstm}). For input data, we concatenate the dynamic, static, and adverse event series elements per patient per time step, so an input from time $t$ is defined as $x_t^p = y^p_t \mdoubleplus D^p_t \mdoubleplus S^p$, where $\mdoubleplus$ denotes vector concatenation and $x_t \in \mathbb{R}^{m}$ where $m = |\mathcal{F}_s| + |\mathcal{F}_d| + 1 $. For a fixed input window size of $n$, a stacked RNN architecture passes hidden state outputs from the layer $l$, $h_{t,l}$, as inputs to the next layer $x_{t,l+1}$, where $1\leq t \leq n$. For our RNN-BOF with $Z$ stacked layers, the output corresponds with the hidden state vector represented by $h_{n,Z}$, which is fully connected to a single neuron with sigmoid activation for learning the binary probability. This differs from other multivariate time series forecasters \cite{BandaraForecasting} where the last layer is instead fully connected to the output window for the learning of many continuous values. This also differs from stacked RNNs as implemented for binary classification tasks, where all hidden states from the final layer, $\{h_{t,l}\}^{l\leq Z}$, would be flattened and connected to a dense layer before again being connected to a single dense neuron with sigmoid activation. The output of the final neuron is defined by the equation: $P(y^p_{n+1} = 1) = \sigma (\mathbf{w}_d  \cdot  h_{n,Z} + \beta)$, where $\mathbf{w}_d \in \mathbb{R}^d$ is the learned weight vector, and $\beta \in \mathbb{R}$ is the learned bias and $P(y^p_{n+1} = 1)$ is the learned probability that the next day in the adverse event series of a given patient belongs to the positive class.

We make use of two techniques to deal with potential overfitting on small datasets. Dropout layers \cite{dropout} are added after each LSTM layer (Fig. ~\ref{fig:rnnlstm}) which set values in the hidden layer output vectors, $h_{t,}$, to 0 at a fixed rate, and L2 kernel regularisation which has been used in RNNs as a tool to reduce overfitting \cite{l21} calculated by $L2 = \frac{\lambda}{2} \sum w_b^2$, where $w_b$ is a specific hidden weight during back propagation and $\lambda$ is a learned hyper-parameter.

\subsection{Sliding window global training}\label{slidingWindow}
\begin{figure}[htb]
\centering
\includegraphics[width = 8.85cm]{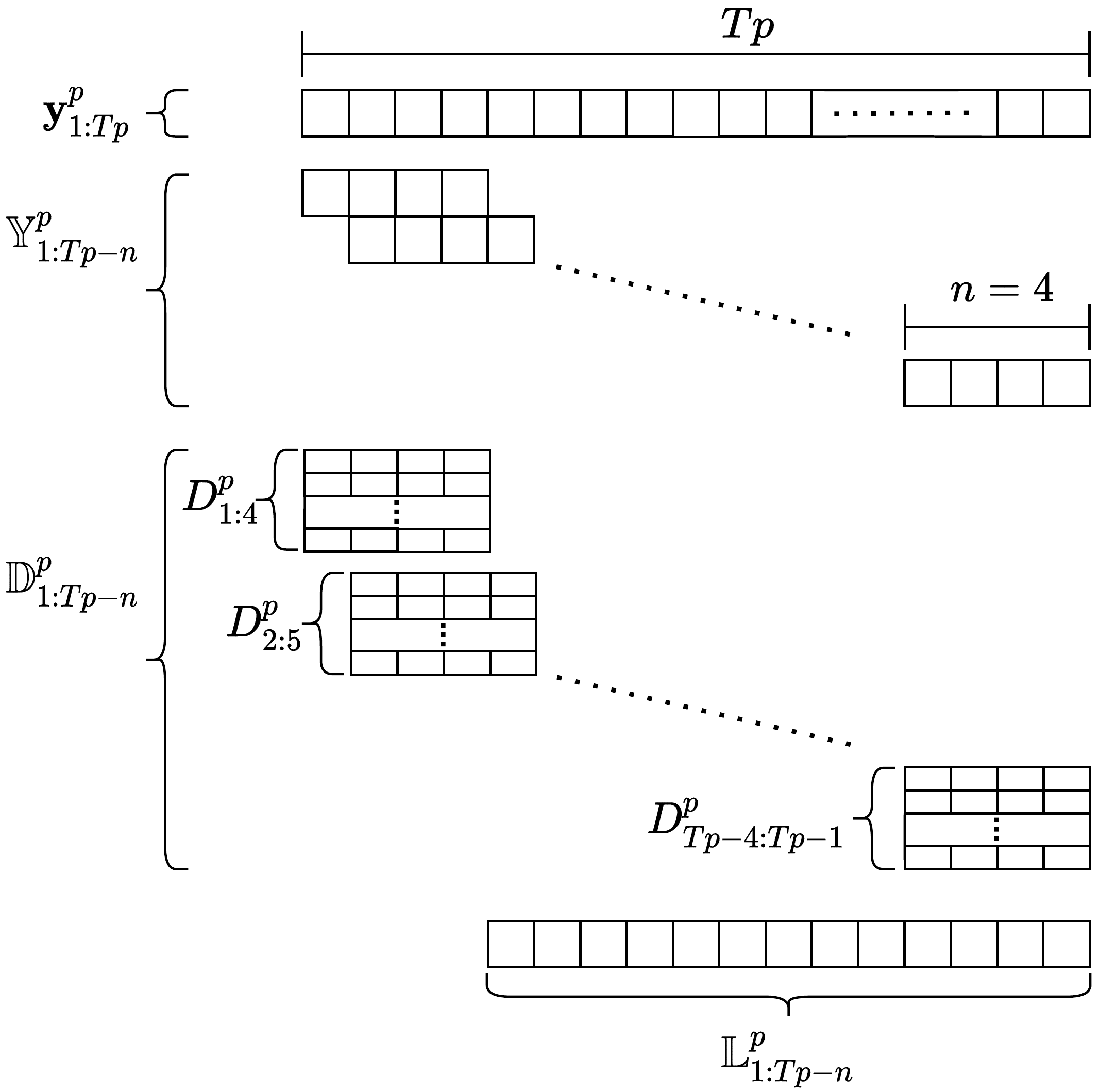}
\caption{The adverse event series windows, $\mathbb{Y}^p_{1:Tp-n}$, dynamic feature series windows, $\mathbb{D}^p_{1:Tp-n}$, and labels, $\mathbb{L}^p_{1:Tp-n}$, associated with patient $i$, for window length $n = 4$.}
\label{fig:windowSet}
\end{figure}

We train our model using a supervised learning sliding window training scheme, again similar to that described by Bandara et al.~\cite{BandaraForecasting}, but where the forecasting point is only the next value in the $\mathbf{y}^p_t$ series, which we will refer to as the label associated with any given input window. We employ this training scheme as it completely maximises use of the relatively small dataset. For each patient, $p \in P$, we define their ordered set of $n$-length dynamic feature windows as $\mathbb{D}^p_{1:Tp-n} = (D^p_{1:1+(n-1)},D^p_{2:2+(n-1)} ,..., D^p_{Tp-n:Tp-1})$, similarly, their adverse event windows are defined as $\mathbb{Y}^p_{1:Tp-n}  ={ (\mathbf{y}^p_{1:1+(n-1)},\mathbf{y}^p_{2:2+(n-1)}, ... , \mathbf{y}^p_{Tp-n:Tp-1})}$, and static covariates are still the set $S^p$. Given windows, w, for patient p, $D^p_{w:w+n-1}$ and $\mathbf{y}^p_{w:w+n-1}$ have associated label $y^p_{w+n}$ where $w \in \mathbb{N} | w \leq (Tp - n)$. We can then define the label vector associated with $\mathbb{D}^p_{1:Tp-n}$, $\mathbb{Y}^p_{1:Tp-n}$ and $S^p$ as $\mathbb{L}^p_{1:Tp-n} = (y^p_n, y^p_{n+1}, ..., y^p_{Tp}) = \mathbf{y}^p_{n:Tp}$ (Fig. ~\ref{fig:windowSet}). 

\subsection{Training and evaluation sets}

For training and evaluation purposes we split each individual patient's windows sequentially into the train and evaluation sets instead of using a held-out patient group. This represents the real world case where a forecaster is implemented on a group of series to predict future values of those series and is common practice in global time series forecasting evaluation~\cite{BandaraSales, BandaraForecasting}. We define this for a 80-20 train test split by taking the last $\rho^p$ windows per patient, where $\rho^p = \left \lfloor {0.2 \cdot |Tp|} \right \rfloor $ so that across different window lengths $n$, the set of test windows will match exactly. The train and test window sets are then defined as $\mathbb{D}_{train} = \{ \mathbb{D}^p_{1:\rho^p-1}\}^{p\in P}$, $\mathbb{Y}_{train} = \{ \mathbb{Y}^p_{1:\rho^p-1 }  \}^{p\in P}$, with associated labels, $\mathbb{L}_{train} = \{ \mathbb{L}^p_{1:\rho^p-1 }  \}^{p\in P}$ and $\mathbb{D}_{test} = \{ \mathbb{D}^p_{\rho^p:Tp}\}^{p\in P}$, $\mathbb{Y}_{test} = \{ \mathbb{Y}^p_{\rho^p:Tp}  \}^{p\in P}$, with associated labels, $\mathbb{L}_{test} = \{ \mathbb{L}^p_{\rho^p: Tp }  \}^{p\in P}$. Thus, we maintain temporal integrity as no window labels are shared across the sets.

\subsection{Loss function and learning algorithm}\label{learning}
Unlike time series forecasting problems with continuous outputs where loss functions such as L2 or L1 are used to learn model weights, we use weighted binary cross entropy batch loss $wbl_B$:

\begin{equation}
    \frac{1}{N_b}\sum_{i = 1}^{N_b}-{y}^p_{n+1}\cdot log(\hat{y}^p_{n+1})  \cdot \gamma_{p} - (1-{y}^p_{n+1})\cdot log(1-\hat{y}^p_{n+1}) \cdot \gamma_{n}
\end{equation}

\noindent where $N_b$ is the size of the batch, and weight, $\gamma_p$ is the ratio of positive to negative days in the training set, and $\gamma_n$ the opposite. For each batch, the model's weights are updated via the Adam learning algorithm. This learning scheme closely represents that of a standard supervised learning classification problem.

\section{Experiments}

In this section we detail our experimental procedure for evaluating the performance of RNN-BOF as compared to benchmarks.

\subsection{Dataset}

We evaluate RNN-BOF on the Thomas Embling Hospital risk assessment dataset (TEH dataset) provided by Swinburne University of Technology\footnote[3]{Ethics approval granted by Swinburne University of Technology (Project \#1258)}. The dataset contains 40 static demographic and general clinical features, 73 features that vary daily or over longer periods, and daily recordings indicating the presence of an aggressive incident. The features include binary representations of general clinical data such as `history of drug use', all components of START and DASA assessments, ordinal summations of both measures as well as general demographic data such as age.

\subsection{Pre-Processing}

We categorised the following three types of missing values found in the dataset: all patients had entries where the collection of dynamic features, psychometric assessments and presence of aggressive incidents spanned more than 24 hours; some patients had components of or full psychometric measures missing for some days; and a small subset were missing the label for presence of aggressive events. The frequency and length of the data entries lasting more than 24 hours were greater at the start and end of a given patient's admission, with one or two long spans of daily assessments. We defined these long frequent assessment periods as a sequence of at least 30 days, starting and ending with 5 daily entries, and with an average of 1.1 days or less between entries over the entire period. As the contents of assessments lasting more than 24 hours were still accurate, no imputation or duplication was performed but a flag was added. For the dynamic missing values, nearest value linear interpolation was used and another flag was added, while patients that were missing values in the aggression series were dropped entirely. We performed z-score standardisation calculated per feature across the entire population's training set. The cleaned frequent assessment periods were then windowed and split 80-20 into train and test window sets as described in Section~\ref{slidingWindow}. Descriptive statistics for the dataset after pre-processing can be seen in Table~\ref{tab:descStats}.

\begin{table}[htb]
\centering

\caption{Processed TEH dataset descriptive statistics comparing the train and testing sets with patient count of $|P| = 83$ .}
\footnotesize
\begin{tabular}{l|l}

Description & Value\\ \hline
Windows in train and test set respectively & 9343 / 2296 \\
Windows with aggression labels in train and test set  & 545 / 122 \\
Avg.  of series length                            & 140.23 \\
Std. Dev.  of series length                        & 105.78 \\
Avg. of positive labeled windows per patient & 8.03  \\
Std. Dev. of positive labeled windows per patient &  20.20 \\
Static / dynamic feature counts                       & 40 / 74   \\ \hline

\end{tabular}
\label{tab:descStats}
\end{table}

\subsection{Evaluation metric} \label{metric}

Previous works in the clinical psychology domain have commonly evaluated the predictive performance of their psychometric instruments using the ROC curve and its respective area \cite{ParghiAssessing,OrruMachine}. The ROC curve is not attenuated by imbalanced datasets such as ours and as such optimistically evaluates performance on the minority class, masking poorly performing models \cite{imb1}. We instead use the Precision Recall Gain curve (PRG) \cite{PRG}, a variation on the Precision Recall (PR) curve which has been shown to be a stronger predictive evaluator than the ROC on imbalanced datasets \cite{imb1}. While representing the same relationship between precision and recall, the Area Under the PRG Curve (AUC-PRG) is valid under cross-validation averaging and the curve does not require non-linear interpolation between points. The AUC-PRG is defined by the following equation:

    \begin{equation}
    \textit{AUC-PRG} = \int_{0}^{1} \textit{precG d recG}
    \end{equation}
    
\noindent Here, FP, FN, TP are false positive, false negative, and true positive, respectively, and 
     \begin{equation}
    \textit{precG} = 1-\frac{\pi}{1-\pi}\frac{FP}{TP} \: , \: \textit{recG} = 1-\frac{\pi}{1-\pi}\frac{FN}{TP}
    \end{equation}

\subsection{Benchmark models}

We chose benchmark models that have seen use in clinical psychology machine learning papers as probabilistic or scored predictive classifiers. This list includes Random Forests (RF) \cite{ParghiAssessing,OrruMachine}, Support Vector Machines (SVM) \cite{IsaacQuantitative}, Feed Forward Neural Networks (FFNN) \cite{OrruMachine} and logistic regression \cite{ParghiAssessing,OrruMachine}. We also included Light Gradient Boosted Machines (LGBM) due to their probabilistic output capability and for being strong global time series forecasters \cite{HewamalageGlobal}. All benchmark models were trained using the same global sliding window training scheme across all patients as described in our modelling approach (Section~\ref{slidingWindow}). The benchmarks were trained and tested using the same input window lengths of 10 and 20 days and therefore are also acting as binary times series forecasters. An additional one day length input window was also tested, representing the traditional classification approach described in Section~\ref{Learning}.

For psychometric instrument benchmarks we used the two measures collected in the TEH dataset; the DASA assessment \cite{OgloffDASA} and the vulnerability component of START assessment \cite{STARTOriginal}. The DASA total score is calculated by summing seven individual binary scored components, each representing the presence of a risk indicating behavior. The START vulnerability assessment contains 20 components, each scored 0, 1 or 2 on a scale representing vulnerability that is summed to give a score between 0 and 40.

\subsection{Overall procedure}\label{trainingSection}

The complete set of windows were split sequentially 80-20 per patient, then pooled as the train and test set respectively (Section~\ref{slidingWindow}). For each model's hyper-parameter tuning we used 5-fold expanding sequential cross-validation on the pool of patients' training windows \cite{HyndmanForecasting}, with folds instead of single instances and pooling across a population. The cross-validation procedure splits the training windows per patient into six sequential and equal size window pools. A validation model is then trained on the first pool and validated on the second, and a second validation model is then trained on the first two pools and validated on the third and so on. This still maintains the temporal integrity described by Hyndman and Athanasopoulos~\cite{HyndmanForecasting} as no labels are shared between train and validation folds. The AUC-PRG is averaged across validation folds to inform a Bayesian hyper-parameter tuning algorithm (namely the Hyperopt algorithm, for 100 iterations). After parameter selection, each model is trained on the full training set \cite{BandaraSales} with forecasts then performed on the held-out test windows for final performance evaluation.

\subsection{Hyper-parameters}
We chose the hyper-parameters and their ranges through existing literature \cite{dropout,l21,BandaraSales} and manual exploration on the training set cross-validation folds. The tuned parameters and their respective ranges identified for RNN-BOF were: number of hidden nodes per layer, 10 - 100; number of stacked layers, 1-4; dropout percent, 10\% - 50\%; L2 regularisation, 0.000001 - 0.001; and number of epochs, 10 - 100. We used a fixed batch size of 512 and found that too high of a learning rate led to especially poor performance so we fixed it at 0.0001. We note that all benchmarks except the FFNN showed worse performance with the weighted loss despite the imbalanced nature of the output windows, so standard binary cross entropy loss was used.

\section{Results and discussion}

In this section we provide a comparison of all tested models and psychometrics instruments, as well as interpretation and discussion of these results. The test set contains a total of $2296$ windows with $5.31\%$ having a positive label (and $5.83\%$ in the training window set, Table~\ref{tab:descStats}). As described in Section~\ref{slidingWindow}, regardless of input window length, windows contain the same forecast labels which allows us to make like-to-like comparisons. We calculated confidence intervals for the AUC-PRG using stratified moving block bootstrapping (the requirement of normality was validated using the Kolmogorov-Smirnov Test) \cite{movingBlock,strata} on the complete set of test window predictions.  We compare the performance of models directly by calculating the bootstrapped distribution of differences in AUC-PRG across the same samples, using the null hypothesis that our model performs equal or worse we can  get a $p$-value indicating a significant increase of AUC-PRG. In Table~\ref{tab:ptable} we provide all models' AUC-PRG with associated 95\% confidence intervals, along with the difference between AUC-PRG of RNN-BOF with 10 day input windows (RNN-BOF10) and all other models where $p$-values indicate a bootstrapped significant difference in area (B=2000). RNN-BOF10 is the best performing model with an AUC of $0.976$ on the test set and an increase in AUC-PRG against all benchmarks at the $p < 0.05$ significance level. 

\begin{table}[htb]
\centering
\caption{AUC-PRG for all models' predictions on the test dataset sorted from best to worst. Difference in AUC-PRG between RNN-BOF10 and all other models, with $p$-values indicating the significance of increase of AUC-PRG. }

\resizebox{8.88cm}{!}{
\begin{tabular}{l|l|l|l|l}
Model$^{\mathrm{a}}$  & AUC-PRG & \multicolumn{1}{p{1.4cm}|}{AUC-PRG CI 95\%}  & \multicolumn{1}{p{1.3cm}|}{AUC-PRG Dif.} & $p$-value$^{\mathrm{b}}$ \\ \hline
RNN-BOF10     & \textbf{0.9761} & {[}0.9535,0.9852{]} & -                & -              \\
RNN-BOF20     & 0.9664 & {[}0.9428,0.9815{]} & 0.0097             & $p = 0.1280$           \\  \hline
FFNN20      & 0.9602 & {[}0.9336,0.9781{]} & 0.0159             & $p = 0.0345$          \\
FFNN10      & 0.9599 &  {[}0.9324,0.9777{]} & 0.0162             & $p = 0.0485$          \\
FFNN1       & 0.9598 &  {[}0.9265,0.9771{]}  & 0.0164            & $p = 0.0175$          \\
RF1        & 0.9566 & {[}0.9312,0.9745{]} & 0.0195             & $p = 0.0210$           \\
RF10       & 0.9559 & {[}0.9241,0.9754{]} & 0.0202             & $p = 0.0255$          \\
LGBM20     & 0.9518 & {[}0.9191,0.9734{]} & 0.0243             & $p = 0.0085$          \\
RF20       & 0.9489 & {[}0.9150,0.9703{]} & 0.0272             & $p = 0.0045$          \\
LGBM10     & 0.9477 & {[}0.9180,0.9694{]} & 0.0284             & $p = 0.0115$          \\
LR1        & 0.9441 & {[}0.9088,0.9705{]} & 0.0320             & $p = 0.0060$           \\
LGBM1       & 0.9394 & {[}0.6599,0.7872{]} & 0.0367             & $p = 0.0030$           \\
SVM20      & 0.9072 & {[}0.8335,0.9457{]} & 0.0689             & $p \leq 0.0005 $           \\
SVM10      & 0.878 &  {[}0.8016,0.9263{]} & 0.0981             & $p \leq 0.0005 $            \\
SVM1       & 0.8733 & {[}0.7781,0.9233{]}  & 0.1028             & $p \leq 0.0005 $            \\
DASA       & 0.8679 & {[}0.7733,0.9203{]} & 0.1082             & $p \leq 0.0005  $           \\
LR10       & 0.8542 & {[}0.7595,0.9187{]} & 0.1219             & $p \leq 0.0005  $           \\
LR20       & 0.7895 & {[}0.6828,0.8694{]} & 0.1866             & $p \leq 0.0005  $            \\
STARTVuln  & 0.7264 & {[}0.6599,0.7872{]} & 0.2497             & $p \leq 0.0005 $            \\ \hline

\multicolumn{5}{p{9cm}}{$^{\mathrm{a}}$Number after model name indicates input window length.} \\
\multicolumn{5}{p{9cm}}{$^{\mathrm{b}}$Calculated on stratified moving block bootstrapped samples with B = 2000 (a horizontal line indicates the significance threshold of 0.05).}
\end{tabular}
}
\label{tab:ptable}
\end{table}

Unlike for ROC curves where the main diagonal, $y = x$, is the baseline random classifier (AUC-ROC = 0.5), for PRG the equivalent diagonal, $\textit{precG} = \textit{recG}-1$, represents the always positive classifier \cite{PRG}. Subsequently, for PRG analysis of a dataset with low amounts of positive labels the values of $\textit{precG}$ and $\textit{recG}$ will be higher due to a worse performing baseline. The drawback of this can be that differences at high precision and recall levels on an imbalanced dataset are harder to determine and indicate a bigger difference than they would for the equivalent ROC and PR curves. 

\begin{figure}[htb]
\centering
\includegraphics[scale=.29]{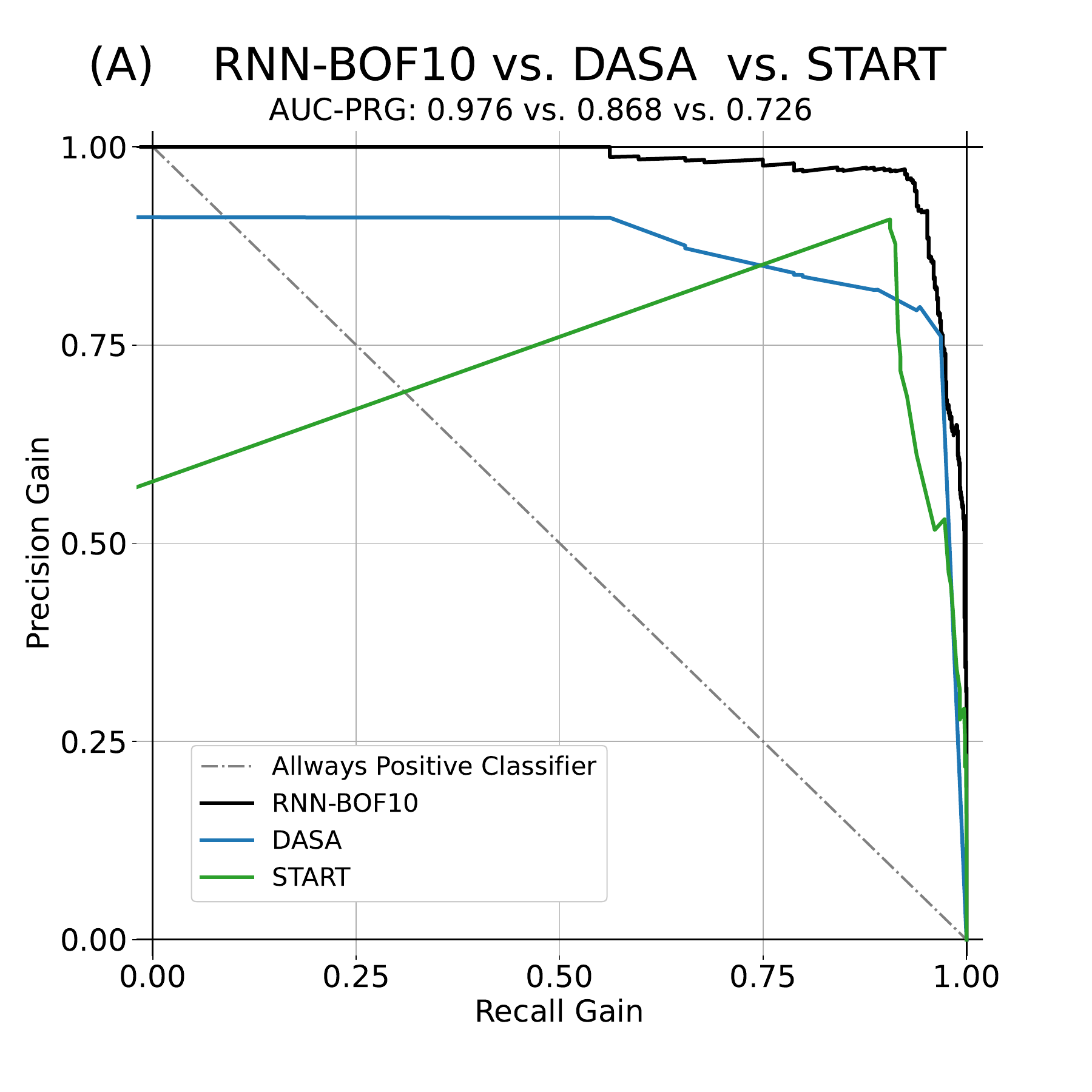}
\includegraphics[scale=.29]{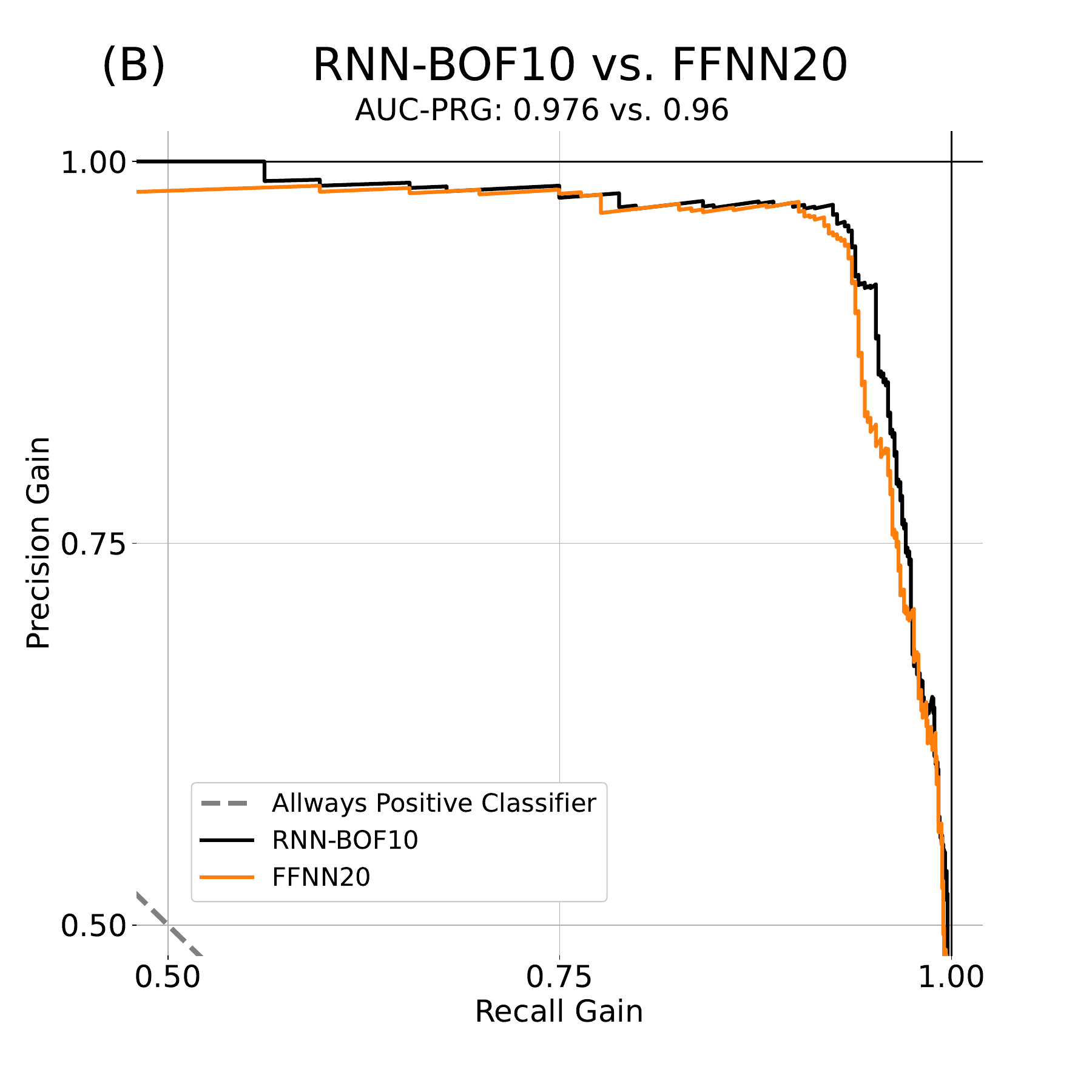}

\caption{AUC-PRG plots comparing (A) RNN-BOF with 10 day input windows (RNN-BOF10) against DASA and START psychometrics and (B) the top-rightmost region of comparison between RNN-BOF10 and a FFNN with 20 day input windows (FFNN20). Note, different scales used in panels A and B. }
\label{fig:4plot}
\end{figure}

We provide three comparison PRG curves to better understand where our model gains its performance increase. In Fig. ~\ref{fig:4plot}A we compare RNN-BOF10 with the psychometric instruments START and DASA where we can see that the majority of the difference in area comes from the top-rightmost region, and high precision gain values. In practical terms the higher precision gain values indicate our model can determine a higher number of true positives from false positives for the same amount of false negatives. The points in the top-rightmost area represent the common threshold calibration use case where precision is roughly equal to recall and points on the line parallel to the always positive classifier and that are closest to $\textit{precG = 1}$ and $\textit{recG = 1}$ represent peak calibrated $F_1$ scores for the model \cite{PRG}. The difference between RNNBOF10 and the best performing benchmark machine learning model, FFNN with 20 day input windows (FFNN20) (Fig. ~\ref{fig:4plot}B), are much smaller, with only slight gaps at high precision gain levels and in the top-rightmost region, although these regions still result in a significant performance improvement.

\section{Conclusion}

Our research builds upon previous machine learning works in both the field of clinical risk assessment and global time series forecasting. We presented a novel approach to clinical risk assessment, specifically for predicting instances of inpatient aggression that leveraged daily longitudinal data collection. We achieved this using RNN-BOF, which unlike previous global multivariable approaches is capable of probabilistic forecasts on binary time series. Our model uses a modified multivariable Recurrent Neural Network with a single probabilistic output value for the estimation of next day patient risk, differing architecturally from RNNs as used for multivariable classification and forecasting tasks. The model was globally trained on a real world dataset across sequentially constructed windows of all patients consisting of: the target binary time series, representing aggressive incidents; covariate series, representing psychometric measurements and dynamic clinical features; and static clinical and demographic features. Unlike the benchmark models, RNN-BOF had an explicit understanding of the sequential nature of the data, allowing for the potential to learn temporal relations that may exist within a patient's historical data. Our model outperformed both the currently used psychometric tools and accepted machine learning methods on a dataset comprised of many clinical variables recorded daily from many patients.

Our study is not without limitations: the dataset contained only 83 patients from a single population with some patients having as little as 30 days of usable sequential data. Further works would need to be performed before statements can be made about the general efficacy of our methodology as a clinical risk assessment tool. Additionally, we focused our analysis of predictive performance on the PRG curve and we believe it to be the most appropriate metric. However, for different use cases, different metrics and analysis may be necessary.

With the popularity and performance of global time series models, forecasters are no longer limited to homogeneous series or single series modelling \cite{MonteroPrinciples}. We hope this work serves to indicate that tasks which traditionally used classification techniques, and where frequent data collection occurs, may benefit from this global time series approach which can explicitly learn temporal relationships.

\bibliographystyle{IEEEtran}
\bibliography{bibt}

\end{document}